\documentclass[runningheads]{llncs}
\usepackage[T1]{fontenc}
\usepackage{graphicx}
\begin{document}
\title{EngineAD: A Real-World Vehicle Engine Anomaly Detection Dataset}
%
%
\author{Hadi Hojjati\inst{1,2}\orcidID{0000-0001-7271-0587} \and
Christopher Roth\inst{3}\and
Rory Woods\inst{3}\and Ken Sills\inst{3}\and Narges Armanfard\inst{1,2}\thanks{Corresponding author.}\orcidID{0000-0002-5880-906X}}
\authorrunning{H. Hojjati et al.}
%
\institute{Department of Electrical and Computer Engineering, McGill University, CA \and
MILA - Quebec AI Institute, CA \and
Preteckt Inc., Hamilton, ON, Canada\\
\email{narges.armanfard@mcgill.ca}}
\maketitle              
\begin{abstract}
The progress of Anomaly Detection (AD) in safety-critical domains, such as transportation, is severely constrained by the lack of large-scale, real-world benchmarks. To address this, we introduce EngineAD, a novel, multivariate dataset comprising high-resolution sensor telemetry collected from a fleet of 25 commercial vehicles over a six-month period. Unlike synthetic datasets, EngineAD features authentic operational data labeled with expert annotations, distinguishing normal states from subtle indicators of incipient engine faults. We preprocess the data into $300$-timestep segments of $8$ principal components and establish an initial benchmark using nine diverse one-class anomaly detection models. Our experiments reveal significant performance variability across the vehicle fleet, underscoring the challenge of cross-vehicle generalization. Furthermore, our findings corroborate recent literature, showing that simple classical methods (e.g., K-Means and One-Class SVM) are often highly competitive with, or superior to, deep learning approaches in this segment-based evaluation. By publicly releasing EngineAD, we aim to provide a realistic, challenging resource for developing robust and field-deployable anomaly detection and anomaly prediction solutions for the automotive industry.

\end{abstract}
\section{Introduction}

Anomaly detection is the task of identifying unusual patterns in data that deviate from the system’s expected behavior, often signaling failures, security breaches, or other rare but critical events \cite{HOJJATI2024106106}. Within the transportation sector, anomaly detection plays a crucial role in ensuring the safety, reliability, and efficiency of vehicles. Modern vehicles generate vast amounts of multivariate sensor data from subsystems such as engines, braking systems, and environmental sensors. Anomalies in these signals can indicate incipient faults, hazardous driving conditions, or potential safety-critical failures \cite{ecml}. Detecting such events in a timely manner enables predictive maintenance, prevents costly breakdowns, and enhances overall road safety. Unlike synthetic testbeds or simulated industrial datasets, anomalies in vehicle telemetry are naturally occurring, often subtle, and context-dependent, posing unique challenges for TSAD \cite{11025380}. These anomalies can manifest as isolated point deviations, such as abrupt fluctuations in engine temperature, or as longer temporal sequences, such as gradual degradation in fuel efficiency or vibration signatures \cite{11136008}.

Despite substantial advances in anomaly detection algorithms, the progress of this field remains constrained by the lack of large-scale, high-quality, and domain-relevant datasets. Most existing benchmarks either rely on synthetic anomalies injected into otherwise normal data streams or are limited in scope and complexity. While such datasets provide a useful starting point, they fail to capture the variability, noise, and context-specific patterns that characterize anomalies in real-world systems \cite{9835419}. 

Furthermore, most existing benchmark datasets suffer from flawed design, mislabeled sequences, and biased evaluation metrics. For example, benchmarks often include trivial anomalies, artificially injected outliers, or unrealistic anomaly ratios, raising concerns about whether reported improvements reflect genuine progress or artifacts of the dataset design. These issues are exacerbated in sequential data, where temporal dependencies introduce additional challenges. Moreover, current evaluation practices can introduce biases that inadvertently favor noisy predictions. Together, these shortcomings hinder fair comparisons across methods and obscure our understanding of true advances in the field \cite{liu2024the}.

To address the limitations of existing resources, we introduce a new multivariate dataset collected from a fleet of commercial vehicles over a six-month period. The dataset includes one-second interval recordings from thirteen engine-related sensors and is complemented by expert annotations of each vehicle’s operational state. These annotations, derived from technician assessments, indicate whether the vehicle was operating normally or exhibiting early signs of potential engine faults, providing reliable ground-truth labels for anomaly detection. Unlike many existing datasets, our collection captures authentic operational conditions, reflecting the variability of real-world environments and the temporal dynamics of long-term system behavior.

The contributions of this work are threefold. First, we release a domain-specific, real-world dataset for anomaly detection \b{and anomaly prediction} in commercial vehicle monitoring. Second, we provide high-resolution, temporally aligned sensor data with expert diagnostic labels, supporting analysis of both transient and gradual anomalies. Third, by making this dataset publicly available, we aim to establish a benchmark for systematic evaluation, foster reproducibility, and stimulate the development of methods capable of addressing the inherent complexities of vehicle anomaly detection.

\section{Related Works}

Research in anomaly detection can be situated along three interrelated dimensions: anomaly taxonomies, methodological paradigms, and benchmarking efforts.

\subsection{Anomaly Taxonomies}

Anomalies in vehicles can be categorized into three types \cite{reviewpang}: (i) point anomalies, where individual observations deviate from expected values, such as sudden spikes in pressure; (ii) contextual anomalies, which depend on temporal or environmental context, such as unusual braking sequences; and (iii) sequence anomalies, involving subsequences that deviate from learned temporal patterns, such as irregular engine vibration patterns. Effective benchmarks and detection methods must account for this diversity to ensure applicability in real-world automotive settings.

\subsection{Methodological Paradigms}
Traditional anomaly detection methods are often based on statistical assumptions (e.g., ARIMA, PCA-based detectors), distance measures (e.g., Matrix Profile, kNN-based methods), or density-based models (e.g., LOF, Isolation Forest) \cite{boniol2024divetimeseriesanomalydetection}. With advances in deep learning, neural methods such as autoencoders, recurrent neural networks, convolutional architectures, and transformers have been widely explored. Models like OmniAnomaly, TranAD, and Anomaly Transformer exemplify prediction-based and representation-based approaches that achieve strong results on existing benchmarks \cite{deepsurvey}. More recently, the introduction of foundation models has shifted attention toward general-purpose pretraining. In the context of sequential data, models such as, MOMENT, Chronos, and Lag-Llama demonstrate strong few-shot and zero-shot capabilities but face challenges in detecting long-duration sequence anomalies and carry risks of data contamination from large-scale pretraining \cite{kottapalli2025foundationmodelstimeseries}.

\subsection{Benchmarking Efforts}
Benchmark datasets have played a central role in shaping anomaly detection research. However, very few publicly available datasets contain engine anomalies for vehicles \cite{s23115013}. In contrast, more resources exist for other types of sequential data. Early datasets such as Yahoo, Numenta, and UCR provided accessible testbeds but were criticized for containing trivial anomalies, flawed labeling, or unrealistic anomaly ratios \cite{liu2024the}. Prior works argued that such benchmarks created an “illusion of progress.” More recent efforts have attempted to overcome these shortcomings \cite{9835419}. The most comprehensive to date, TSB-AD, introduced over 1,000 curated time series and systematically addressed dataset flaws and evaluation biases. Their findings challenged prevailing assumptions by showing that simpler statistical approaches often outperform advanced neural architectures, while foundation models excel primarily at detecting point anomalies \cite{liu2024the}. These insights highlight both the limited state of knowledge in anomaly detection for temporal data and the critical need for appropriately labeled datasets.

While these benchmarks mark significant progress, they remain limited in domain coverage. Most existing datasets originate from generic industrial processes, synthetic anomaly injections, or controlled testbeds, with few capturing anomalies from safety-critical domains such as transportation. Our dataset addresses this gap by providing authentic, anomaly-labeled data from real vehicles, reflecting naturally occurring faults and abnormal conditions. Unlike synthetic datasets, it captures the complexity of real-world anomaly distributions, including rare events and context-dependent sequences. By positioning our dataset within the broader benchmarking landscape, we aim to support fairer evaluation of anomaly detection methods and broaden the scope of benchmarks available for the automotive domain.

\section{Method}

\subsection{Data Recording}
\begin{figure}
    \centering
    \includegraphics[width=0.8\linewidth]{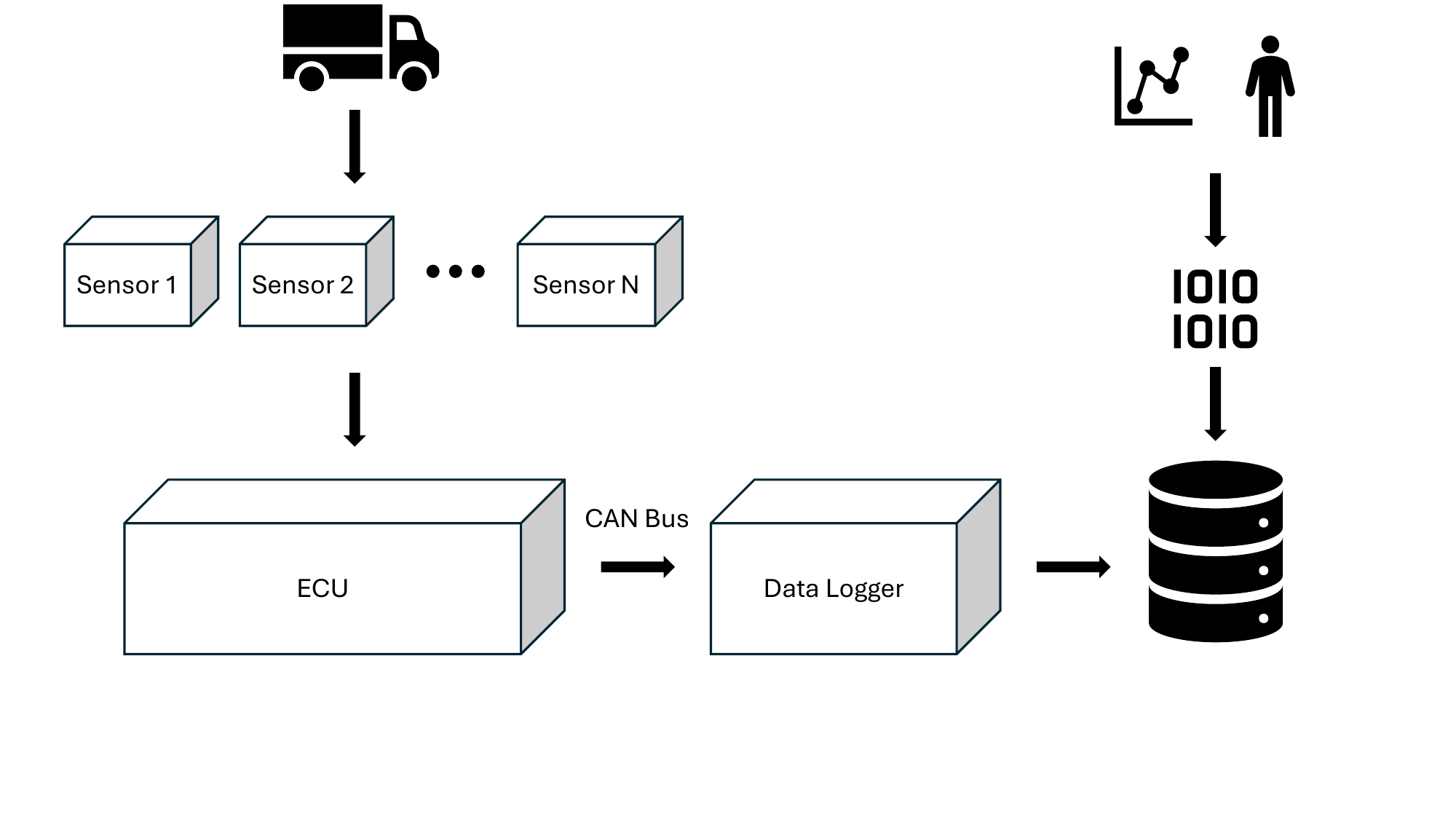}
    \caption{Overview of Data Recording Process. Sensor signals from the vehicles were transmitted via the CAN bus and captured using a proprietary logging device connected to the network. Following data collection, a team of technicians systematically analyzed the sensor traces and maintenance information to assign ground-truth labels.}
    \label{fig:datacuration}
\end{figure}

\begin{figure*}
    \centering
    \includegraphics[width=1\linewidth]{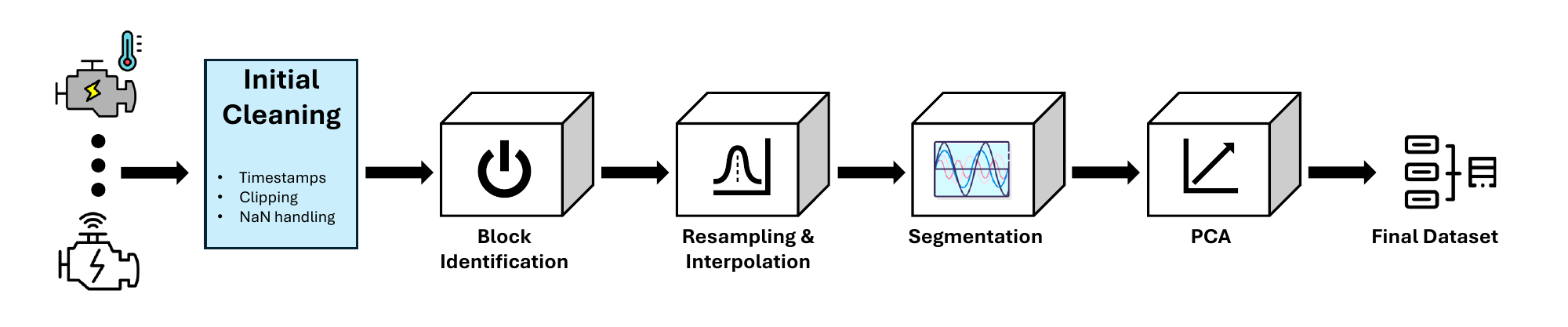}
    \caption{Preprocessing pipeline for vehicle sensor data. Thirteen sensors were selected with technician input. Data were cleaned, engine-on periods identified, resampled to 1Hz, and segmented into non-overlapping five-minute windows. Incomplete or missing segments were removed, and each vehicle’s labeled segments were saved for analysis.}
    \label{fig:flow}
\end{figure*}

The dataset comprises time-series sensor readings collected from a fleet of 25 trucks. All vehicles are of the same model and manufacturer, ensuring an identical configuration of sensors across the fleet. Data were recorded continuously over a six-month period, spanning August to February. The signals originate from a range of engine-related sensors, including measurements of pressure, temperature, fuel rate, and rotational speed, among others. These raw sensor values are transmitted through the vehicle’s Controller Area Network (CAN) bus, which acts as the internal communication backbone for electronic control units (ECUs). Each sensor reading is encoded into CAN frames with a unique identifier, timestamp, and payload representing the measurement. The CAN bus streams were captured using dedicated logging devices, enabling synchronized collection of multi-sensor data.

To provide ground-truth labels, the recording company employed a team of technicians who systematically analyzed both the sensor traces and maintenance reports. Through this process, they annotated operational states, distinguishing between normal conditions and patterns indicative of potential defects that may precede engine failure. These expert annotations transform the raw DATA streams into a labeled dataset suitable for anomaly detection, while retaining the complexity and heterogeneity of real-world vehicle operation.

\subsection{Preprocessing}

In total, thirteen sensors were employed, each monitoring a distinct engine parameter. The sensors were selected in consultation with experienced vehicle technicians to ensure relevance to engine health and potential failure modes. The raw data were stored in a database and queried for the target vehicles and sensors. After sorting by timestamp, values outside known operational ranges were clipped, and known erroneous readings were replaced with NaN. Rows where all sensor readings were missing were dropped, and duplicate timestamps were resolved by averaging.

To isolate periods of active operation, engine-on blocks were identified using the engine RPM, splitting continuous running periods at gaps longer than 30 seconds. Only these engine-on periods were retained to ensure segments reflected uninterrupted vehicle operation. Data were then resampled to a uniform one-second frequency, with missing values filled via linear interpolation. Labels indicating normal or abnormal operation were carefully aligned with the preprocessed data.

The time series were divided into non-overlapping segments of length $L=300$ (five minutes). Shorter segments at the end of each block were discarded, and any segments containing missing sensor readings were removed to maintain data quality.

Finally, each vehicle’s fully processed dataset was saved as a pickle file, containing a list of clean, labeled segments ready for downstream analysis. This approach ensures that the resulting dataset accurately captures meaningful operational patterns while minimizing noise and inconsistencies.

\subsection{Dataset Annonymization}
To protect the identity of individual vehicles and prepare the dataset for public release, a mapping system was implemented. The original alphanumeric vehicle identifiers, extracted from the filenames, were replaced with simple, sequential, and generic identifiers (e.g., \textit{truck 1}, \textit{truck 2}, etc.).

The primary method used for sensor data anonymization and dimensionality reduction was Principal Component Analysis (PCA). To ensure a consistent and objective transformation, both data normalization and the PCA model were fitted across the entire vehicle fleet. This approach is preferable to training separate models for each vehicle, as it establishes a single, fleet-wide baseline for what constitutes a normal sensor value and the principal sources of variation across all vehicles. The PCA model was configured with $n_{components} = 0.95$, retaining the minimum number of components required to explain 95\% of the total variance in the dataset. This step effectively reduced the dimensionality of the feature space while preserving the vast majority of its information. After running PCA with the aforementioned criteria, 8 principal components were saved for each data point. This procedure ensures that the original data cannot be reconstructed, thereby safeguarding the company’s proprietary information, while still providing practitioners with transformed data that preserves the patterns necessary for effective anomaly detection

\subsection{Dataset Access}

The dataset is available upon request.
\footnote{Visit: https://github.com/Armanfard-Lab/EngineAD for information on how to access the dataset.} 
The final output is a directory containing a collection of pickle files, with each file corresponding to a single anonymized vehicle (e.g., \texttt{truck\_1.pickle}). Each file contains a list of dataframes, where each dataframe represents a fixed-length time series. Each dataframe consists of the principal components, the anomaly label ($0 =$ normal, $1 =$ anomaly), and the anonymized vehicle identifier.

\section{Experiments}

\begin{table*}
    \centering
    \begin{tabular}{|c|c|c|c|c|c|c|c|c|c|}
    \hline
         Truck ID & KMeans & IF &OCSVM &LOF &AE &COPOD &HBOS &SOD &DeepSVDD \\
         \hline
         1 & 0.88 & 0.85 & 0.87 & 0.87 & 0.80 & 0.86 & 0.85 & 0.84 & 0.85 \\
         2 & 0.43 & 0.46 & 0.43 & 0.44 & 0.50 & 0.43 & 0.44 & 0.47 & 0.47 \\
         3 & 0.49 & 0.49 & 0.50 & 0.50 & 0.52 & 0.50 & 0.50 & 0.50 & 0.50 \\
         4 & 0.76 & 0.75 & 0.74 & 0.74 & 0.71 & 0.76 & 0.76 & 0.73 & 0.76 \\
         5 & 0.72 & 0.72 & 0.72 & 0.72 & 0.71 & 0.71 & 0.71 & 0.71 & 0.72 \\
         6 & 0.71 & 0.72 & 0.70 & 0.66 & 0.67 & 0.72 & 0.72 & 0.68 & 0.70 \\
         7 & 0.78 & 0.76 & 0.79 & 0.79 & 0.65 & 0.76 & 0.76 & 0.77 & 0.76 \\
         8 & 0.36 & 0.39 & 0.38 & 0.37 & 0.46 & 0.38 & 0.40 & 0.41 & 0.40 \\
         9 & 0.77 & 0.76 & 0.77 & 0.78 & 0.75 & 0.75 & 0.75 & 0.76 & 0.76 \\
         10 & 0.41 & 0.46 & 0.44 & 0.38 & 0.53 & 0.44 & 0.45 & 0.45 & 0.46 \\
         11 & 0.67 & 0.66 & 0.68 & 0.67 & 0.67 & 0.66 & 0.66 & 0.67 & 0.66 \\
         12 & 0.59 & 0.59 & 0.59 & 0.59 & 0.56 & 0.59 & 0.59 & 0.60 & 0.58 \\
         13 & 0.68 & 0.69 & 0.69 & 0.69 & 0.67 & 0.67 & 0.67 & 0.69 & 0.67 \\
         14 & 0.55 & 0.56 & 0.58 & 0.57 & 0.57 & 0.56 & 0.56 & 0.57 & 0.57 \\
         15 & 0.48 & 0.52 & 0.50 & 0.47 & 0.56 & 0.50 & 0.50 & 0.53 & 0.51 \\
         16 & 0.70 & 0.69 & 0.69 & 0.70 & 0.66 & 0.69 & 0.68 & 0.68 & 0.69 \\
         17 & 0.75 & 0.75 & 0.75 & 0.75 & 0.75 & 0.74 & 0.74 & 0.75 & 0.74\\
         18 & 0.69 & 0.69 & 0.69 & 0.67 & 0.68 & 0.69 & 0.69 & 0.68 & 0.69 \\
         19 & 0.93 & 0.92 & 0.93 & 0.93 & 0.90 & 0.91 & 0.91 & 0.90 & 0.90 \\
         20 & 0.37 & 0.48 & 0.41 & 0.35 & 0.53 & 0.42 & 0.44 & 0.47 & 0.46 \\
         21 & 0.70 & 0.69 & 0.69 & 0.71 & 0.66 & 0.69 & 0.69 & 0.69 & 0.69 \\
         22 & 0.75 & 0.74 & 0.75 & 0.76 & 0.73 & 0.74 & 0.74 & 0.74 & 0.74 \\
         23 & 0.84 & 0.82 & 0.84 & 0.83 & 0.79 & 0.83 & 0.82 & 0.81 & 0.82 \\
         24 & 0.79 & 0.76 & 0.78 & 0.79 & 0.76 & 0.77 & 0.77 & 0.77 & 0.77 \\
         25 & 0.21 & 0.25 & 0.26 & 0.23 & 0.39 & 0.26 & 0.26 & 0.30 & 0.26 \\
         \hline
         \textit{Average} & \textit{0.64}&\textit{0.65}&\textit{0.65}&\textit{0.64}&\textit{0.65}&\textit{0.64}&\textit{0.64}&\textit{0.65}&\textit{0.65}\\

        \hline
    \end{tabular}
    \vspace{6pt}
    \caption{Segment-Based Anomaly F1-Scores Across the EngineAD Fleet.
Performance measured by the F1-score for nine one-class anomaly detection models. Each result represents the performance on a single vehicle, where models were trained on normal segments and tested on a held-out set containing both normal and anomaly segments.}
    \label{tab:results}
\end{table*}

To establish a systematic performance baseline and empirically quantify the complexity of the EngineAD dataset, we conducted an extensive evaluation using a diverse suite of anomaly detection algorithms. These experiments serve to benchmark current capabilities and highlight the specific challenges posed by real-world engine data.

\subsection{Evaluation Protocol}

The evaluation was designed as a segment-based one-class anomaly detection task. Given the fixed-length preprocessing step, each multivariate time-series segment (comprising $300$ time steps across $8$ principal components) was treated as a single high-dimensional data instance by flattening it into a feature vector. This methodology enables the application of classical and deep learning-based anomaly detectors by representing the entire sequence's profile.

To accurately simulate the real-world scenario where labeled anomaly data is scarce, the protocol adhered to a one-class learning paradigm, executed individually for each of the anonymized vehicles. The models were trained exclusively on $70\%$ of the segments definitively labeled as normal. The test set comprised the remaining $30\%$ of the normal segments, concatenated with all available anomaly segments. This balanced test set composition ensures that the evaluation is sensitive to both false positives (misidentifying normal operations as anomalous) and false negatives (failing to detect true anomalies).

Prior to model training, all input features were standardized using a standard scaler instance fitted solely on the training data. For threshold-based scoring methods (e.g., K-Means and Autoencoder), the anomaly boundary was determined by setting the threshold at the 95th percentile of the anomaly scores (distance or reconstruction error) observed on the training set. This common practice aims to control the false positive rate on the known healthy data. Performance was measured using the F1-score calculated specifically for the anomaly class, as this metric provides a robust assessment of a detector's efficacy in highly imbalanced and critical tasks.

\subsection{Anomaly Detection Models}

We selected nine representative models spanning four major algorithmic paradigms to ensure a comprehensive baseline study:

\begin{itemize}
    \item \textbf{Statistical and Density-Based Methods}:
    \begin{itemize}
        \item \textbf{HBOS (Histogram-based Outlier Score)}: A fast, univariate method that assumes feature independence and estimates probability densities using histograms \cite{Goldstein2012HBOS}.
        \item \textbf{COPOD (Copula-Based Outlier Detection)}: A sophisticated statistical approach that leverages empirical copula functions to model feature dependencies robustly \cite{copod}.
        \item \textbf{SOD (Subspace Outlier Detection)}: Identifies anomalies by analyzing the local data distribution within various randomly selected subspaces \cite{sod}.
    \end{itemize}

    \item \textbf{Distance and Clustering Methods}:
    \begin{itemize}
        \item \textbf{KMeans Distance (KMeans)}: Models the normal class using a single cluster center; anomaly score is the Euclidean distance to this center \cite{Duda2000}.
        \item \textbf{LOF (Local Outlier Factor)}: A density-based method that quantifies how isolated a data instance is with respect to its local neighborhood \cite{Breunig2000LOF}.
    \end{itemize}
    
    \item \textbf{Ensemble and Boundary-Based Methods}:
    \begin{itemize}
        \item \textbf{Isolation Forest (IF)}: An ensemble method that explicitly isolates anomalies using binary search trees \cite{IF}.
        \item \textbf{One-Class SVM (OCSVM)}: A kernel-based method that learns a maximal margin hyperplane to separate the majority of the data from the origin in a high-dimensional feature space \cite{Schoelkopf1999SVMNovelty}.
    \end{itemize}

    \item \textbf{Deep Learning Methods}:
    \begin{itemize}
        \item \textbf{Autoencoder (AE)}: A neural network trained to minimize the reconstruction error of normal input. Anomalies are detected via high reconstruction error. The network employed a fully-connected encoder-decoder structure: $2400 \rightarrow 1024 \rightarrow 128 \rightarrow 1024 \rightarrow 2400$ \cite{zhou2017anomaly}.
        \item \textbf{Deep SVDD (Deep Support Vector Data Description)}: Maps data using a neural network into a feature space where the smallest hypersphere enclosing the normal data is learned, minimizing the distance of normal points to the sphere's center \cite{Ruff2018DeepOCC}.
    \end{itemize}
\end{itemize}

\subsection{Results and Discussion}

The F1-scores for the anomaly class for all models across the 25 vehicles are summarized in Table \ref{tab:results}.

The most striking finding is the pervasive inter-vehicle performance variability. F1-scores for a single model, such as KMeans, range from highly effective (e.g., $0.93$ for Truck 19) to near-random performance (e.g., $0.21$ for Truck 25). This high variance suggests that the characteristics of engine anomalies are highly non-uniform across the fleet, demanding anomaly detection methods that are robust to significant distributional shifts in the underlying healthy data and failure modes. The inability to achieve high F1-scores on certain vehicles indicates that their anomalies are exceptionally subtle, highly contextual, or closely integrated into the spectrum of normal operating conditions.

Contrary to expectations often placed on modern deep learning models, simpler, non-parametric methods—specifically KMeans Distance and One-Class SVM—demonstrate superior or highly competitive performance against the deep learning-based Autoencoder and DeepSVDD across a majority of the vehicles. This implies that for many anomalies in this dataset, the pattern of deviation from the norm is substantial enough to be captured by global distance or boundary-based measures in the flattened feature space. This observation aligns with recent findings that classical anomaly detection methods can be as effective as, or in some cases superior to, deep learning architectures on contemporary time series benchmarks \cite{liu2024the}.

However, deep learning models are not without merit. In challenging cases where classical methods struggle significantly (e.g., Truck 25, where the Autoencoder achieved the highest F1-score of $0.39$), the capacity of deep models to learn complex, non-linear representations of the normal operational manifold becomes crucial.

In summary, the experiments confirm that EngineAD is a realistic and challenging benchmark. The necessity for high-performing, generalized models that can overcome the observed high vehicle-specific variability remains a significant open research problem.

\section{Conclusion}

In this work, we introduced EngineAD, a novel, large-scale, and domain-specific dataset designed to advance research in engine anomaly detection within safety-critical transportation systems. Derived from $25$ commercial vehicle engines and featuring over six months of continuous, high-resolution sensor telemetry, EngineAD addresses the critical scarcity of non-synthetic, expertly annotated benchmarks in the automotive field. The expert-driven labeling process ensures reliable ground-truth annotations, distinguishing between normal operations and early indicators of engine faults.

By publicly releasing EngineAD, we aim to establish a rigorous, reproducible benchmark that facilitates the fair comparison of future anomaly detection methodologies and stimulates the development of robust, field-deployable solutions for predictive vehicle maintenance. Moving forward, primary research should focus on developing methods that explicitly handle the cross-vehicle data shift through domain adaptation, and on enhancing model interpretability to provide actionable diagnostic insights by linking anomaly scores back to the original sensor space.

\begin{credits}
\subsubsection{\ackname} The authors gratefully acknowledge the financial support provided by the Natural Sciences and Engineering Research Council of Canada (NSERC). The authors also thank Preteckt, Inc. for generously providing the data, as well as their time and expert insights, which were invaluable to the success of this work.
\end{credits}
%
%
%
\bibliographystyle{splncs04}
\bibliography{refs}




\end{document}